\documentclass[letterpaper]{article} 
\usepackage{aaai25}  
\usepackage{times}  
\usepackage{helvet}  
\usepackage{courier}  
\usepackage[hyphens]{url}  
\usepackage{graphicx} 
\usepackage{natbib}  
\usepackage{caption} 
\usepackage{algorithm}
\usepackage{algorithmic}

\usepackage{newfloat}
\usepackage{listings}
\DeclareCaptionStyle{ruled}{labelfont=normalfont,labelsep=colon,strut=off} 
\floatstyle{ruled}
\newfloat{listing}{tb}{lst}{}
\floatname{listing}{Listing}

\usepackage[switch]{lineno} 
\usepackage{xcolor}
\usepackage{cite}
\usepackage{multirow}
\usepackage{subfigure}
\usepackage{xspace}
\usepackage{dsfont}
\usepackage{epstopdf}
\usepackage{booktabs}
\usepackage{amsmath}
\usepackage{amsthm}
\usepackage{amsfonts}
\usepackage{bm}
\usepackage{adjustbox}
\usepackage{appendix}

\newtheorem{mydef}{Definition}
\newcommand{\bitem}[1]{\noindent$\bullet$ \textbf{#1}}
\newcommand{\paratitle}[1]{\noindent\textbf{#1}}
\newcommand{\ie}{\emph{i.e.,}\xspace}

\newcommand{\eg}{\emph{e.g.,}\xspace}

\newcommand{\name}{POI-Enhancer\xspace}
\newcommand{\figureautorefname}{Fig.}
\newcommand{\tableautorefname}{Tab.}

\title{POI-Enhancer: An LLM-based Semantic Enhancement Framework \\ for POI Representation Learning}
\author{
    Jiawei Cheng\textsuperscript{\rm 1,2},
    Jingyuan Wang\textsuperscript{\rm 1,3,4,}\footnote{Corresponding authors.},
    Yichuan Zhang\textsuperscript{\rm 1},\\
    Jiahao Ji\textsuperscript{\rm 1},
    Yuanshao Zhu\textsuperscript{\rm 2},
    Zhibo Zhang\textsuperscript{\rm 1},
    Xiangyu Zhao\textsuperscript{\rm 2,}\footnotemark[1]
}
\affiliations{

    \textsuperscript{\rm 1}SKLCCSE, School of Computer Science and Engineering, Beihang University, Beijing, China\\
    \textsuperscript{\rm 2}Department of Data Science, City University of Hong Kong, Hong Kong, China\\
    \textsuperscript{\rm 3}MIIT Key Laboratory of Data Intelligence and Management, Beihang University, Beijing, China\\
    \textsuperscript{\rm 4}School of Economics and Management, Beihang University, Beijing, China\\
    \{JarvisC, jywang, tadshi, jiahaoji, hljzhangzhibo\}@buaa.edu.cn, yaso.zhu@my.cityu.edu.hk, xy.zhao@cityu.edu.hk
}

\begin{document}

\maketitle

\begin{abstract}
POI representation learning plays a crucial role in handling tasks related to user mobility data. Recent studies have shown that enriching POI representations with multimodal information can significantly enhance their task performance. 
Previously, the textual information incorporated into POI representations typically involved only POI categories or check-in content, leading to relatively weak textual features in existing methods. 
In contrast, large language models (LLMs) trained on extensive text data have been found to possess rich textual knowledge.
However leveraging such knowledge to enhance POI representation learning presents two key challenges: first, how to extract POI-related knowledge from LLMs effectively, and second, how to integrate the extracted information to enhance POI representations.
To address these challenges, we propose POI-Enhancer, a portable framework that leverages LLMs to improve POI representations produced by classic POI learning models. We first design three specialized prompts to extract semantic information from LLMs efficiently. Then, the Dual Feature Alignment module enhances the quality of the extracted information, while the Semantic Feature Fusion module preserves its integrity. The Cross Attention Fusion module then fully adaptively integrates such high-quality information into POI representations and Multi-View Contrastive Learning further injects human-understandable semantic information into these representations. Extensive experiments on three real-world datasets demonstrate the effectiveness of our framework, showing significant improvements across all baseline representations.
\end{abstract}

%
\begin{links}
    \link{Code}{https://github.com/Applied-Machine-Learning-Lab/POI-Enhancer}
    \link{Extended version}{https://arxiv.org/abs/2502.10038}
\end{links}

\section{Introduction}\label{sec:intro}

With the advancement of smart city technology~\cite{cityshield, DGeye,infer} and the widespread adoption of smart devices, the volume of location-based mobile data, such as POI (Points of Interest) check-in data and user trajectory data, has surged~\cite{ding2018ultraman,nipsdataset}. Predicting user destinations~\cite{zhao2020go}, forecasting visit flow~\cite{song2020spatial}, and similar tasks~\cite{hx} have become key research focuses. In tackling these difficulties, POI representation learning, which can be trained via self-supervised methods and utilized across various tasks like traffic forecasting~\cite{fullbayesian, STDEN,energyfield} and trajectory predction~\cite{trajrepr, Astar, CD-CNN}, stands as a particularly meaningful and promising direction.

To enhance the diversity of information within POI representation vectors and achieve superior performance in complex downstream tasks, researchers are exploring the integration of various information beyond basic geographic data. For example, they incorporated user preference data~\cite{ppr} and visual information~\cite{cityfm} into POI representations. Although related textual information, such as POI categories (e.g. restaurants and hotels) and check-in content on social media like Twitter, provides some insights into the social functions and other aspects of POIs, the semantic richness and depth of these data are limited. When compared to the vast amount of descriptive information available on the internet regarding POIs, these data sources fall short in both content richness and coverage. 
In recent years, large language models (LLMs) trained on extensive volumes of internet data have been applied across numerous fields, demonstrating remarkable capabilities, particularly in the domain of spatial-temporal data~\cite{urbangpt}. Although LLMs have proven beneficial in addressing challenges in this area, leveraging LLMs to enhance POI representation presents two specific challenges.

The first challenge lies in effectively \textbf{extracting the geographical knowledge within LLMs}. A common idea~\cite{xval} is to provide LLMs with prompts related to geographic information and then obtain text output. However, LLMs have limitations in handling numerical input, and for representation learning, we need vectors that are versatile across tasks, which makes this method not suitable. Some studies~\cite{gatgpt,stllm} have also experimented with feeding extracted spatial-temporal features to a partially or fully frozen LLM, using the LLM as the backbone to solve specific problems. But these works are typically tailored to a single spatial-temporal task and extract information specific to that task only. However, POI representation learning aims to capture versatile information to address diverse tasks. Clearly, task-specific extraction is insufficient for this requirement.

The second challenge is how to effectively \textbf{integrate the extracted textual information into POI representation} for enhancement. Since the information extracted by LLMs is versatile, combining these diverse aspects information with the POI representation is difficult. Most researchers ~\cite{ppr} adopted the approach of one-hot encoding the corresponding POI category features and then concatenating them with the representation vectors, which overlooks the interactions between features. For example, the POI category and time are related: a restaurant's lunch hours and lunch break times exhibit different visitor flow patterns. This limits the ability to exploit the richness of semantic information to enhance POI representations.

To address the challenges, we propose a POI representation enhancement framework, called \name, which is designed to leverage textual information in LLM to strengthen embedding vectors. Specifically, to better utilize LLMs for extracting textual features of POIs, we develop unique prompts to separately extract features related to POI addresses, visit patterns, and surrounding environments. Following this, we design the Dual Feature Alignment module to leverage the relationships between textual features, enabling the acquisition of higher-quality semantic information. The Semantic Feature Fusion module is specifically designed to ensure the preservation of high-quality semantic information. Then, to fully integrate the extracted information with the representation vectors, we introduce the Cross Attention Fusion module based on the attention mechanism. Finally,  we incorporate Multi-View Contrastive Learning to further inject human-understandable semantic information into POI representations to enhance its capability of capturing real-world patterns. 

We summarize our main contributions as follows:
\begin{itemize}
    \item To the best of our knowledge, \name is the first to introduce LLM-based textual knowledge to enhance POI representations of POI learning models. We demonstrate that leveraging knowledge from LLMs is crucial for boosting the performance of POI embedding models.
    
    \item We design three kinds of specialized prompts to thoroughly extract textual information from LLMs, and employ a cross-attention mechanism to adaptively integrate these information into POI representations. We also introduce temporal, spatial, and functional contrastive learning to refine the POI representations.

    \item We conducted extensive experiments on three public real-world datasets across various downstream tasks. The results demonstrate that our approach significantly enhances the performance of POI embedding methods.
\end{itemize}

\section{Preliminaries}\label{sec:pre}
\begin{mydef}[\textbf{Point of Interest (POI)}]
A POI is a specific geographic location with some basic attributes $p=(id, pn, c, lon, lat)$, where $id$ indicates index, $pn$ means name of POI, $c$ denotes category, $lon$ and $lat$ represent longitude and latitude coordinates respectively. Besides, each POI has some extra attributes such as visit pattern, address, and surrounding environment. 
An example of the attributes of a POI in New York City is provided in \tableautorefname~\ref{tab:poi_example}.
\end{mydef}

\begin{mydef}[\textbf{Check-in Record}]
A check-in record is a triplet $r=(u,p,t)$ which means a user $u$ visits POI $p$ at time $t$. A user's movement behavior over a period of time can be modeled by a sequence of check-in records, which we define as a Check-in Record Sequence. It can be represented by $R = \{r_1,r_2,...,r_L\}$, where the check-in records are arranged in the order of time sequence and $L$ is the length of the Check-in Record Sequence. 
We also denote the set of all users' check-in record sequences as $S$.
\end{mydef}

\begin{mydef}[\textbf{POI Representation}]
Given the set of all POIs $P=\{p_1,~p_2,~\ldots,~p_N\}$, where n is the number of the set, a mapping function $f$ is used to generate a fixed vector representation $E_{p_i} = f(p_i)$ for each POI. 
\end{mydef}

\noindent \textbf{Problem Statement}. Given a POI Representation function $\mathcal{F}$, POIs set $P=\{p_1,~p_2,~\ldots,~p_N\}$ and other related data \eg check-in record sequences $S$, with the aid of LLM, the aim of our framework is to learn a function $g$ that enhance the capability of the function $F$, \ie $E_{p_i} = g(\mathcal{F} (p_i)), \ E_{p_i} \in \mathbb{R}^d$, where $d$ is a uniform dimension. 

\begin{table}[ht]
\centering
    \resizebox{0.7\columnwidth}{!}{%
        \begin{tabular}{rccc}
            \toprule
            Attribute & Value \\
            \midrule
            POI ID &  22337 \\ 
            Name &  New York Stock Exchange \\ 
            Longitude & 74.011154 \\ 
            Latitude & 40.706806 \\
            Category & Stock Exchange \\
            Street Name  & Wall Street\\
            House Number & 11 \\
            Postal Code & 10005 \\
            Surrounding & Office, Building and Road \\
            Visit Pattern & Between 6 am and 9 am, Weekday \\
            \bottomrule
        \end{tabular}
        }
        \caption{An Example of the POI attributes.} 
    \label{tab:poi_example}
\end{table}

\section{Methodology}\label{sec:method}

\begin{figure*}[t]
    \centering
    \includegraphics[width=\textwidth]{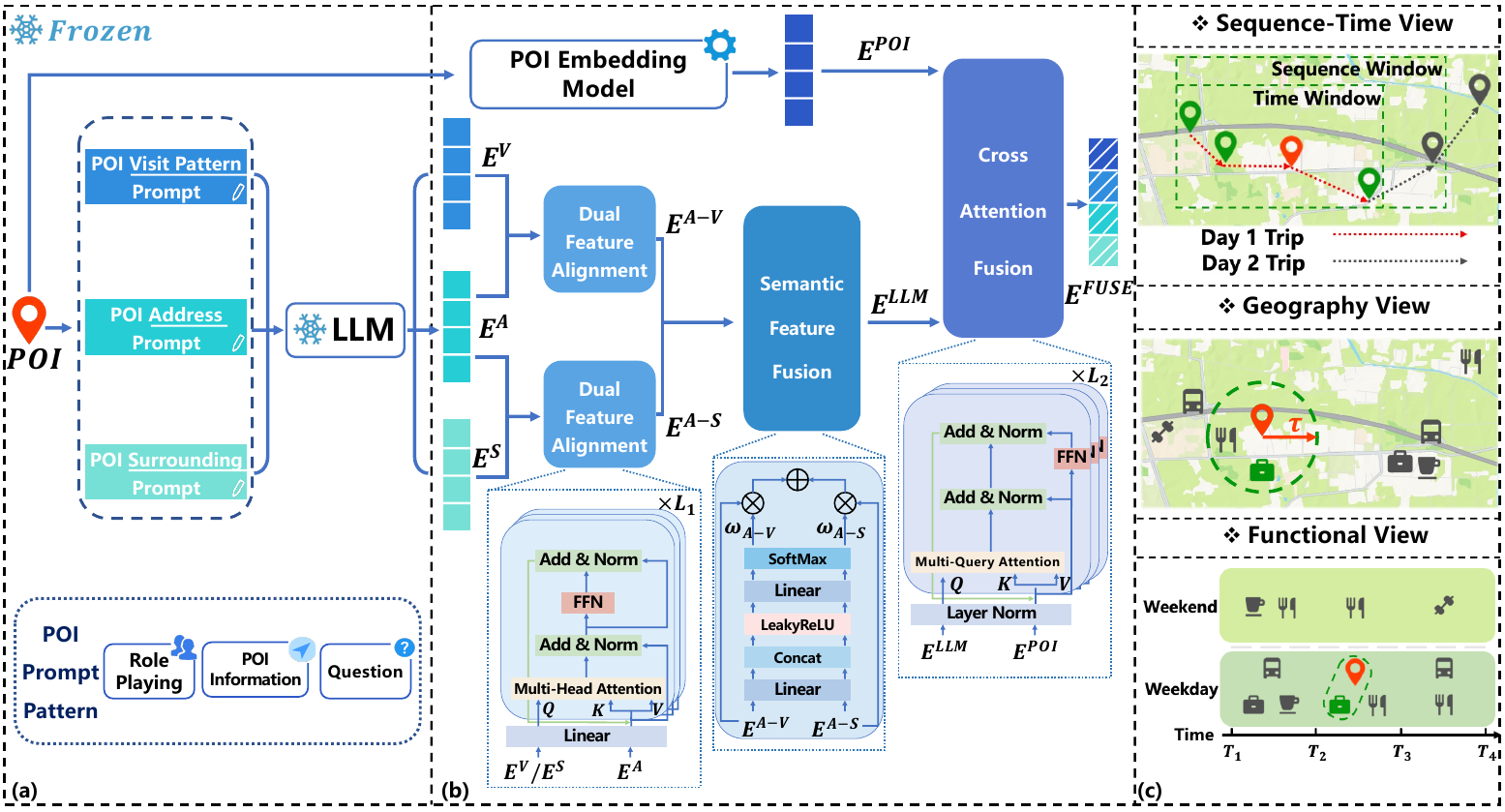} 
    \caption{(a): Prompt Generation and Feature Extraction are used to obtain prompts and get textual features from the LLM. (b): Embedding Enhancement is designed to enhance POI embeddings by leveraging textual features. (c): Multi-View Contrastive Learning enables the sampling of more diverse positive and negative examples during training.}
    \label{fig:framework and prompt}
\end{figure*}

This section provides a comprehensive demonstration of the technical details of \name framework and  \figureautorefname~\ref{fig:framework and prompt} presents the overall architecture. In \figureautorefname~\ref{fig:framework and prompt}, part (a) is the Prompt Generation and Feature Extraction phase, where specialized prompts are generated and used to extract relevant semantic information from the LLM. The second phase, Embedding Enhancement, corresponds to part (b), where the extracted information is further refined and integrated with the POI representations to be enhanced. Finally, part (c) represents Multi-View Contrastive Learning, where we designed three sampling strategies to select positive and negative samples for contrastive learning. Besides, to assist LLMs in more accurately capturing POI-related knowledge, we additionally processed and derived three kinds of extra attributes mentioned above. A detailed description of this procedure can be found in the Supplementary Material.


\subsection{Prompt Generation and Feature Extraction}
    \paratitle{Generate prompt}
    Due to the LLM's low sensitivity to numbers, we need to bundle basic attributes like latitude, longitude, and name with extra attributes when inputting them, to help the LLM accurately target the desired POI. Besides, simply stacking various features into a prompt can make it difficult for the LLM to focus on key points and effectively extract information. Hence, the proposed prompt pattern consists of three parts: (1) Role-Playing, (2) POI Information, and (3) the Question. The POI Information part encompasses basic information and extra information, corresponding to the basic and extra attributes, respectively. Firstly, the design purpose of role-playing at the beginning of the prompt is to allow the LLM to fully unleash its knowledge, enabling the LLM to embody a role familiar with geographical knowledge. An attribute header is added in front of the POI information to help the LLM accurately capture the information of input attributes. Next, we generate multiple sentences based on combinations of the basic attributes and three extra attributes. Lastly, inspired by the ~\cite{gurnee2024worldmodel}, we design the question at the end of each prompt about the content to trigger the relevant knowledge.
    Consequently, we generate three types of prompts for each POI $p_i$:POI Visit Pattern Prompt, POI Address Prompt, and POI Surrounding Prompt, denoted as $T^V_{p_{i}}, T^A_{p_{i}}, T^S_{p_{i}}$. An Example of the prompt we generated is shown in \figureautorefname~\ref{fig: prompt}, which is in Supplementary Material.

    \paratitle{Extract from LLM}
    In \name, we input the prompts into the LLM and take the final hidden layer state from the LLM as the semantic feature. It is worth noting that the LLM serves as a frozen encoder when training. So, for a POI $p_i$, the feature extraction process can be denoted as:
    \begin{equation} \small
        \begin{aligned}
             \boldsymbol{E}^V_{p_i} = \mathcal{H}(T^V_{p_{i}}) , \
             \boldsymbol{E}^A_{p_i} = \mathcal{H}(T^A_{p_{i}}) , \
             \boldsymbol{E}^S_{p_i} = \mathcal{H}(T^S_{p_{i}}) ,
        \end{aligned}
    \end{equation}

    \noindent where $\boldsymbol{E}^V_{p_i},\boldsymbol{E}^A_{p_i},\boldsymbol{E}^S_{p_i}\in \mathbb{R}^{D}$ are the corresponding semantic feature of three kinds of prompts, $\mathcal{H}$ is the process of extracting the last hidden state from the LLM, and $D$ is the dimension size of the hidden state vector.

\subsection{Embedding Enhancement}
    \textit{Dual Feature Alignment} leverages the intricate connections between address and visit patterns, as well as between address and surrounding environment to obtain higher-quality semantic features. \textit{Semantic Feature Fusion} uses attention score-weighted merging to ensure the quality of the features when fusing the semantic features into a single semantic vector. Afterward, \textit{Cross Attention Enhancement}, based on the cross-attention method, employs the semantic vector obtained earlier to fully integrate and enhance the POI representations, resulting in the final output vector.
    
    \paratitle{Dual Feature Alignment}
    A POI's address, a key factor of geography information, is closely linked to its visit patterns and surrounding environment. For example, as shown in \tableautorefname ~\ref{tab:poi_example}, the New York Stock Exchange is on Wall Street, a well-known hub of financial firms. People often visit there during daytime working hours, and the surrounding environment mainly consists of office spaces. If we align the address textual feature with the visit pattern and surrounding environment textual features, we can obtain higher-quality textual information.
    Thus, we designed Dual Feature Alignment. First, Given a batch of textual information of $n$ POIs $\{E^V, E^A, E^S\}$, they will be fed into a linear layer to transform them into a hidden space with the same dimension as the POI embedding to be enhanced, denoted as:
     \begin{equation}\small
        \boldsymbol{\tilde{E}}^V=\boldsymbol{W}^{V'} \boldsymbol{E}^V , \
        \boldsymbol{\tilde{E}}^A=\boldsymbol{W}^A \boldsymbol{E}^A , \
        \boldsymbol{\tilde{E}}^S=\boldsymbol{W}^S \boldsymbol{E}^S  ,
    \end{equation}
    \noindent where $\boldsymbol{\tilde{E}}^{V},\boldsymbol{\tilde{E}}^A,\boldsymbol{\tilde{E}}^S \in \mathbb{R}^{n \times d}$, $d$ is the dimension of the hidden space and $\boldsymbol{W}^{V'},\boldsymbol{W}^A,\boldsymbol{W}^S$ are all learnable matrices.

    Next, to leverage the relationships between textual features and obtain higher-quality information, multiple layers of the Transformer encoder are introduced. Each layer consists of multi-head attention ($\mathrm{MHA}$), residual connections, and layer normalization operations ($\mathrm{LN}$) and the number of layers is $L_1$.
    Formally, take the relation between address and visit patterns as an example, given  the vectors $\{\tilde{E}^V,\tilde{E}^A\}$, we computed a $\mathrm{MHA}$ as follows:
    \begin{equation}\small \label{eq:qkv} 
        \begin{aligned}
            \boldsymbol{Q} = \boldsymbol{\tilde{E}}^V \boldsymbol{W}^Q  ,\
            \boldsymbol{K} = \boldsymbol{\tilde{E}}^A \boldsymbol{W}^K  ,\
            \boldsymbol{V} = \boldsymbol{\tilde{E}}^A \boldsymbol{W}^V  ,
        \end{aligned}
    \end{equation}
    \begin{equation}\small
        \begin{aligned}
            head_h = \phi(\frac{\boldsymbol{Q}\boldsymbol{K}^T}{\sqrt{d}})\boldsymbol{V}  ,\\
        \end{aligned}
    \end{equation}
    \begin{equation}
        \begin{aligned}\small \label{eq:mha}
            \mathrm{MHA}(\boldsymbol{\tilde{E}}^V,\boldsymbol{\tilde{E}}^A)=(\Vert^H_{h=1}head_h)\boldsymbol{W}^O  ,
        \end{aligned}
    \end{equation}
    \noindent where $\boldsymbol{W}^Q,\boldsymbol{W}^K,\boldsymbol{W}^V \in \mathbb{R}^{d \times d_h}$ are learnable parameters, $\phi$ is softmax activation function,  $d_h$ is the dimension of a single head. And $\Vert$ is the concatenation operation, $\boldsymbol{W}^O \in \mathbb{R}^{ (d_h \cdot H) \times d}$ is a trainable parameter and $H$ denotes the number of heads.
    The output of the first layer $Z_1'$ is denoted as:
    \begin{equation}\small \label{eq:addnorm}
        \begin{aligned}
           \boldsymbol{Z} = \mathrm{LN}(\boldsymbol{\tilde{E}}^A+\mathrm{MHA} (\boldsymbol{\tilde{E}}^V,\boldsymbol{\tilde{E}}^A)) ,
        \end{aligned}
    \end{equation}
    \begin{equation}\small
        \begin{aligned}
           \boldsymbol{Z}_1' = \mathrm{LN}(\boldsymbol{Z}+\mathrm{FFN}(\boldsymbol{Z})) ,
        \end{aligned}
    \end{equation}
    \noindent where $\mathrm{FFN}$ is a feed-forward neural network. Then, vector $\boldsymbol{Z}_1'$, along with $\boldsymbol{E}^{A}$, will be fed back into the next layer as input, and after repeating this process $L_1-1$ times, the final layer result $\boldsymbol{Z}_{L_1}'$ is the vector $\boldsymbol{E}^{A-V} \in \mathbb{R}^{n \times d}$. It should be noted that $\{Z_k'| k \in [1,L_1]\}$ is transformed into $K$ and $V$, while  $\tilde{E}^A$ is converted into $\boldsymbol{Q}$ in the subsequent repetition process.
    Similarly, to deal with the connection between address and surrounding by replacing $\boldsymbol{\tilde{E}}^V$ with $\boldsymbol{\tilde{E}}^S$ in Formula ~(\ref{eq:qkv}), (\ref{eq:mha}) and (\ref{eq:addnorm}). We can get the output $E^{A-S}$ accordingly.

    \paratitle{Semantic Feature Fusion}
    Considering that visit patterns are related to the surrounding environment, for example, POIs near entertainment venues are mostly accessed on weekends. We got a comprehensive semantic feature by combining the two feature vectors from the previous module into one.
    To integrate two vectors into one while maintaining the quality of the vector, we designed the Semantic Feature Fusion based on a weighted sum method. 
    Accordingly, the computation process can be represented as follows:
    \begin{equation}\small
        \begin{aligned}
            \theta^{A-V} = \boldsymbol{W}_2 \cdot LeakyReLU([\boldsymbol{W}_1 \boldsymbol{E}^{A-V} || \boldsymbol{W}_1  \boldsymbol{E}^{A-S}]), \\ 
            \theta^{A-S} = W_2 \cdot LeakyReLU([\boldsymbol{W}_1 \boldsymbol{E}^{A-S} || \boldsymbol{W}_1  \boldsymbol{E}^{A-V}]),
        \end{aligned}
    \end{equation}
    \noindent where $\theta^{A-V}$ and $\theta^{A-S}$ are the attention scores for $\boldsymbol{E}^{A-V}$ and $\boldsymbol{E}^{A-S}$. $\boldsymbol{W}_1 \in \mathbb{R}^{d \times d'}$ and $\boldsymbol{W}_2 \in \mathbb{R}^{2d' \times 1} $ are used to project the features into the same hidden space and to transform them into attention weights, respectively.  $LeakyReLU$ is an activation function, and $d'$ is the dimension of the latent space.
    After that, a softmax activation is employed to get the normalized weight, and a weighted sum fusion of two semantic features is applied to get the output  $E^{LLM} \in \mathbb{R}^{n \times d}$, which can be represented as: 
    \begin{equation}\small
        \begin{aligned}
        [\omega^{A-V},\omega^{A-S}] = \phi([\theta^{A-V},\theta^{A-S}]) ,
        \end{aligned}
    \end{equation}
    \begin{equation}\small
        \begin{aligned}
            \boldsymbol{E}^{LLM} = \omega^{A-V} \cdot \boldsymbol{E}^{A-V} +  \omega^{A-S} \cdot \boldsymbol{E}^{A-S}.
        \end{aligned}
    \end{equation}

    \paratitle{Cross Attention Fusion}
    Cross-attention techniques have been employed to fully fuse features from different views ~\cite{dafusion}. Hence, inspired by ~\cite{urbanclip}, to enhance other embedding methods by making use of the vector $E^{LLM}$, a Cross Attention Fusion is developed.  

    Here, we also employ a multi-layer transformer encoder architecture but in each layer we use the multi-query attention ~\cite{MultiQueryAttention} plus parallel attention and feed-forward net (PAF) ~\cite{parallelAttention} to combine $E^{LLM}$ and $E^{POI} \in \mathbb{R}^{n \times d}$. The multi-query attention (MQA) is almost the same as the multi-head attention except all heads share the same set of $K$ and $V$, which is proved to be faster with minor quality degradation in the calculation. Additionally,  PAF can be effective in improving the performance of transformer-based models. As shown in \figureautorefname~\ref{fig:framework and prompt}, the first layer of the Cross Attention Fusion can be presented formally as:
    \begin{equation}\small
        \begin{aligned}
            \boldsymbol{X} = \mathrm{LN}(\boldsymbol{E}^{POI}+MQA(\boldsymbol{E}^{LLM},\boldsymbol{E}^{POI})) ,
        \end{aligned}
    \end{equation} 
    \begin{equation}
        \begin{aligned}
            \boldsymbol{X}_1' = \mathrm{LN}(\boldsymbol{X}+\mathrm{FFN}(\boldsymbol{X})) ,
        \end{aligned}
    \end{equation}
    
    Then, the vector $\boldsymbol{X}_1'$ and $\boldsymbol{E}^{POI}$
    will be fed into the next layer, and after repeating this process in $\boldsymbol{L}_2-1$ times, the outcome of the last layer $\boldsymbol{E}_{FUSE}$ is obtained. It is worth to noticed that $\{\boldsymbol{X}_k'|k\in [1,L_2]\}$ is transformed into $\boldsymbol{K}$ and $\boldsymbol{V}$, and $\boldsymbol{E}^{LLM}$ is converted into $\boldsymbol{Q}$ in the following repetition.

\subsection{Multi-View Contrastive Learning}
    Our Multi-View Contrastive Learning approach is designed to enhance the similarity between the anchor POI and positive samples, while simultaneously reducing the similarity with negative samples. This strategy aims to strengthen the robustness and effectiveness of the embedding vector.
    However, Unlike previous works that only use distance as the sampling criterion~\cite{urbanfootnote},  we incorporated temporal, spatial, and functional views into our considerations and designed three sampling strategies. Besides, the formal definitions of the following three sampling strategies are presented in the Supplementary Material.
    
    \paratitle{Sequence-Time Contrastive Learning}
    The visit context of a POI \ie the neighboring check-in records in the check-in record sequence is often considered an important factor. However, if the duration of a check-in record sequence is very long, two adjacent consecutive check-in records may be separated by several days. Considering such neighbors as positive samples will reduce the effectiveness of contrastive learning. Therefore, to avoid this situation, we propose a Sequence-Time sampling strategy. The positive samples are required not only to be neighbors of the check-in record but also to have the same visit date as the anchor sample.

    \paratitle{Geography Contrastive Learning}
    From a spatial perspective, our strategy incorporates both local spatial similarity and category similarity as criteria. Specifically, for a given POI, we define a square area centered around it and consider POIs of the same category in that area as positive samples.

    \paratitle{Functional Contrastive Learning}
    Apart from the two types of contrastive learning mentioned above, we aim to identify groups of POIs that are similar in social function. Therefore, based on the category and visit patterns of POIs, We propose the following principle for selecting positive samples: only POIs that share the same category and visit pattern as the anchor sample are regarded as positive samples.
    \newline
    In summary, based on the above three criteria, we sampled more high-quality positive samples for subsequent contrastive learning training. This approach helps enhance the comprehensive capability of the representation vectors.

    \subsection{Model Training}
    
    Given a POI $p_i$ and a set of all the POIs $P$, we derive its positive set $P_i^+$ through the above strategies. And for each pair in $\{(p_i,p_i^+)|p_i^+\in P_i^+\}$, we will randomly choose $m-2$ negative samples from the negative set $\{p_i^-|p_i^- \in P_i^-, P_i^-= P-p_i-P_i^+\}$,  to form a training batch.
    
    Then we use InfoNCE as the loss for contrastive learning:
    \begin{equation}\small
        \mathcal{L}_{Cont} = -log\frac{e^{\frac{1}{\gamma}sim(p_i, p_i^{+})}}{\sum_{i=0}^{m}e^{\frac{1}{\gamma}sim(p_i, p_i^{-})}}
    \end{equation}
    where $sim(\cdot, \cdot)$ is a similarity measure function,  $\gamma$ is a temperature parameter and $m$ is number of POIs in the batch.

    Apart from this, in order to maintain the similarity between the origin vectors and enhanced vectors, a loss based on cosine similarity is constructed, which can be defined as:
    \begin{equation}\small
        \begin{aligned}
            &\mathcal{L}_{Sim} =  \\
        &\frac{1}{m^2}\sum^m_{i=1}\sum^m_{j=1}|cos(E^{FUSE}_i,E^{FUSE}_j)-cos(E^{POI}_i,E^{POI}_j)|,
        \end{aligned}
    \end{equation}

    \noindent where $cos$ is the cosine similarity function.

    Ultimately, the loss of \name  can be denoted:
    \begin{equation}\small
        \mathcal{L} = \mathcal{L}_{Cont} + \mathcal{L}_{Sim}.
    \end{equation}

\section{Experiments}\label{sec:expt}
\subsection{Experiment Setup}
\paratitle{Datasets}
We conducted experiments on three check-in datasets provided by~\cite{4square}: Foursquare-NY, Foursq-SG, and Foursquare-TKY, sampled from New York, Singapore, and Tokyo, respectively. We remove all POIs with less than 5 check-ins in the dataset and check-in sequences with less than 10 POIs. The statistics of the processed dataset are in Supplementary Material.
Then we shuffled the dataset and split it into a ratio of 2:1:7 for the test set, validation set, and training set. It should be noted that the training set for the POI Recommendation task will also be used as the dataset for sampling in contrastive learning. 

\paratitle{Baselines}
We introduced six baselines in our experiment including Skip-Gram~\cite{mikolov2013efficient}, POI2Vec~\cite{feng2017poi2vec}, Geo-Teaser~\cite{zhao2017geo}, TALE~\cite{wan2021pre}, Hier~\cite{shimizu2020enabling}, and CTLE~\cite{lin2021pre}. The details of the baselines are in the Supplementary Material. LLM-based baselines are also included like Llama2~\cite{llama2}, ChatGLM2~\cite{chatglm}, and GPT-2~\cite{gpt2}.

\paratitle{Downstream Tasks \& Metrics}
To evaluate \name and make a comprehensive comparison, we set up three downstream tasks based on LibCity~\cite{libcity}.

\bitem{POI Recommendation},
Based on a user's historical check-in sequence, the POI Recommendation task aims to predict the next POI the user would visit.      

\bitem{Check-in Sequence Classification},
Given an arbitrary check-in sequence, this task requires the downstream model to detect which user this sequence belongs to.

\bitem{POI Visitor Flow Prediction},
POI visitor flow prediction requires the downstream model to forecast the future volume of visitor flow based on historical visitor data.

In the POI Recommendation task, we use Hit@$k$ as the evaluation metric (value equals  1 if the ground truth is among the top k in the recommendation list, otherwise 0, \ $k=1,~5$). The Check-in Sequence Classification task is evaluated using Accuracy (ACC) and Macro-F1 while the Visitor Flow Prediction task is assessed with Mean Absolute Error (MAE) and Root Mean Square Error (RMSE).

\paratitle{Implementation}
In our framework, we use the Llama-2-7B as the LLM backbone. The complete implementation details are provided in the Supplementary Material.

\begin{table*}[t]
\belowrulesep=0pt
\aboverulesep=0pt
\setlength{\tabcolsep}{1mm}
\centering
\resizebox{\linewidth}{!}{
\begin{tabular}{l|cc|cc|cc|cc|cc|cc|cc|cc|cc}
    \toprule
    \textbf{Task} & \multicolumn{6}{c|}{\textbf{POI Recommendation}} & \multicolumn{6}{c|}{\textbf{Check-in Sequence Classification}} & \multicolumn{6}{c}{\textbf{POI Visitor Flow Prediction}} \\
    \midrule
    Dataset & \multicolumn{2}{c|}{NY} & \multicolumn{2}{c|}{TKY} & \multicolumn{2}{c|}{SG} & \multicolumn{2}{c|}{NY} & \multicolumn{2}{c|}{TKY} & \multicolumn{2}{c|}{SG} & \multicolumn{2}{c|}{NY} & \multicolumn{2}{c|}{TKY} & \multicolumn{2}{c}{SG} \\
    \midrule
    Metric & Hit@1 & Hit@5 & Hit@1 & Hit@5 & Hit@1 & Hit@5 & Acc   & F1    & Acc   & F1    & Acc   & F1    & MAE   & RMSE  & MAE   & RMSE  & MAE   &  RMSE  \\
    \midrule
    Skip-Gram & 6.984  & 17.356  & 15.037  & 33.305  & 9.194  & 23.652  & 48.967  & 0.224  & 59.982  & 0.413  & 43.768  & 0.229  & 0.371  & 0.747  & 0.494  & 0.691  & 0.665  & 0.994  \\
    Skip-Gram+ & 7.610  & 18.032  & \textcolor[rgb]{ .122,  .137,  .161}{15.557 } & \textcolor[rgb]{ .122,  .137,  .161}{34.197 } & \textcolor[rgb]{ .122,  .137,  .161}{10.747 } & \textcolor[rgb]{ .122,  .137,  .161}{24.468 } & 52.151  & 0.251  & 62.936  & 0.438  & \textcolor[rgb]{ .122,  .137,  .161}{47.285 } & 0.255  & 0.336  & 0.514  & 0.492  & 0.668  & 0.621  & 0.890  \\
    \textbf{Impr.} & \textbf{8.96\%} & \textbf{3.89\%} & \textbf{3.46\%} & \textbf{2.68\%} & \textbf{16.89\%} & \textbf{3.45\%} & \textbf{6.5\%} & \textbf{12.07\%} & \textbf{4.92\%} & \textbf{5.97\%} & \textbf{8.04\%} & \textbf{11.37\%} & \textbf{9.43\%} & \textbf{31.14\%} & \textbf{0.47\%} & \textbf{3.31\%} & \textbf{6.66\%} & \textbf{10.45\%} \\
    \midrule
    POI2Vec & 6.417  & 14.684  & 15.195  & 33.214  & 8.828  & 21.729  & 32.057  & 0.113  & 51.499  & 0.331  & 31.736  & 0.139  & 0.343  & 0.574  & 0.531  & 0.764  & 0.625  & 0.918  \\
     POI2Vec+ & 7.851  & 18.353  & 15.800  & 34.768  & 10.630  & 24.030  & 52.151  & 0.245  & 62.358  & 0.438  & 46.521  & 0.264  & 0.326  & 0.492  & 0.490  & 0.696  & 0.602  & 0.868  \\
    \textbf{Impr.} & \textbf{22.35\%} & \textbf{24.99\%} & \textbf{3.98\%} & \textbf{4.68\%} & \textbf{20.41\%} & \textbf{10.59\%} & \textbf{62.68\%} & \textbf{117.35\%} & \textbf{21.09\%} & \textbf{32.39\%} & \textbf{46.59\%} & \textbf{89.32\%} & \textbf{4.78\%} & \textbf{14.27\%} & \textbf{7.8\%} & \textbf{8.99\%} & \textbf{3.66\%} & \textbf{5.36\%} \\
    \midrule
    Geo-Teaser & 6.174  & 15.355  & 14.956  & 33.814  & 8.768  & 22.851  & 38.296  & 0.149  & 54.852  & 0.355  & 39.511  & 0.182  & 0.394  & 0.778  & 0.498  & 0.696  & 0.623  & 0.913  \\
    Geo-Teaser+ & \textcolor[rgb]{ .216,  .235,  .263}{7.116 } & \textcolor[rgb]{ .216,  .235,  .263}{16.657 } & 15.500  & 34.475  & \textcolor[rgb]{ .122,  .137,  .161}{10.122 } & \textcolor[rgb]{ .122,  .137,  .161}{23.532 } & 49.910  & 0.233  & 62.647  & 0.437  & 50.064  & 0.279  & 0.341  & 0.524  & \textcolor[rgb]{ .122,  .137,  .161}{0.483 } & \textcolor[rgb]{ .122,  .137,  .161}{0.669 } & 0.588  & 0.854  \\
    \textbf{Impr.} & \textbf{15.27\%} & \textbf{8.48\%} & \textbf{3.64\%} & \textbf{1.95\%} & \textbf{15.45\%} & \textbf{2.98\%} & \textbf{30.33\%} & \textbf{55.84\%} & \textbf{14.21\%} & \textbf{23.11\%} & \textbf{26.71\%} & \textbf{52.98\%} & \textbf{13.35\%} & \textbf{32.64\%} & \textbf{3.07\%} & \textbf{3.87\%} & \textbf{5.57\%} & \textbf{6.41\%} \\
    \midrule
    TALE  & 6.025  & 13.618  & 13.608  & 30.612  & 7.555  & 19.238  & 33.950  & 0.127  & 51.521  & 0.330  & 33.112  & 0.140  & 0.336  & 0.645  & 0.523  & 0.716  & 0.639  & 0.926  \\
    TALE+ & 6.690  & 15.208  & 14.940  & 33.223  & 8.694  & 20.342  & 50.689  & 0.240  & 63.380  & 0.448  & 47.719  & 0.263  & 0.320  & 0.482  & 0.510  & 0.701  & \textcolor[rgb]{ .122,  .137,  .161}{0.610 } & \textcolor[rgb]{ .122,  .137,  .161}{0.903 } \\
    \textbf{Impr.} & \textbf{11.04\%} & \textbf{11.67\%} & \textbf{9.79\%} & \textbf{8.53\%} & \textbf{15.08\%} & \textbf{5.74\%} & \textbf{49.3\%} & \textbf{88.82\%} & \textbf{23.02\%} & \textbf{35.82\%} & \textbf{44.11\%} & \textbf{87.29\%} & \textbf{4.76\%} & \textbf{25.17\%} & \textbf{2.49\%} & \textbf{2.08\%} & \textbf{4.56\%} & \textbf{2.49\%} \\
    \midrule
    Hier  & 6.982  & 15.631  & 15.120  & 32.091  & 9.181  & 22.174  & 37.436  & 0.143  & 50.189  & 0.316  & 41.269  & 0.196  & 0.361  & 0.584  & 0.536  & 0.733  & 0.634  & 1.000  \\
    Hier+ & 8.009  & 19.197  & 16.187  & 35.715  & \textcolor[rgb]{ .122,  .137,  .161}{10.592 } & \textcolor[rgb]{ .122,  .137,  .161}{24.079 } & 51.893  & 0.254  & 63.380  & 0.441  & 47.795  & 0.258  & 0.313  & 0.483  & \textcolor[rgb]{ .122,  .137,  .161}{0.510 } & \textcolor[rgb]{ .122,  .137,  .161}{0.719 } & 0.574  & 0.804  \\
    \textbf{Impr.} & \textbf{14.72\%} & \textbf{22.81\%} & \textbf{7.06\%} & \textbf{11.29\%} & \textbf{15.37\%} & \textbf{8.59\%} & \textbf{38.62\%} & \textbf{77.33\%} & \textbf{26.28\%} & \textbf{39.59\%} & \textbf{15.81\%} & \textbf{31.5\%} & \textbf{13.09\%} & \textbf{17.33\%} & \textbf{4.88\%} & \textbf{1.91\%} & \textbf{9.39\%} & \textbf{19.58\%} \\
    \midrule
    CTLE  & 6.653  & 14.594  & 14.859  & 31.852  & 8.625  & 20.218  & 40.103  & 0.181  & 55.030  & 0.369  & 41.805  & 0.206  & 0.337  & 0.566  & 0.515  & 0.703  & 0.697  & 1.061  \\
     CTLE+ & \textcolor[rgb]{ .216,  .235,  .263}{7.093 } & \textcolor[rgb]{ .216,  .235,  .263}{17.032 } & \textcolor[rgb]{ .216,  .235,  .263}{15.479 } & 34.138  & \textcolor[rgb]{ .122,  .137,  .161}{10.315 } & \textcolor[rgb]{ .122,  .137,  .161}{24.027 } & 50.430  & 0.234  & 61.848  & 0.434  & 51.440  & 0.287  & 0.291  & 0.456  & \textcolor[rgb]{ .122,  .137,  .161}{0.495 } & \textcolor[rgb]{ .122,  .137,  .161}{0.689 } & 0.610  & 0.892  \\
    \textbf{Impr.} & \textbf{6.61\%} & \textbf{16.71\%} & \textbf{4.18\%} & \textbf{7.18\%} & \textbf{19.59\%} & \textbf{18.84\%} & \textbf{25.75\%} & \textbf{29.59\%} & \textbf{12.39\%} & \textbf{17.36\%} & \textbf{23.05\%} & \textbf{39.62\%} & \textbf{13.65\%} & \textbf{19.44\%} & \textbf{4.01\%} & \textbf{2.02\%} & \textbf{12.4\%} & \textbf{15.92\%} \\
    \bottomrule
    \end{tabular}%
} 
\caption{The overall performance of downstream tasks and (+) means being enhanced by \name. Hit@1, Hit@5 and Acc are in \%, and F1 means macro-F1. For MAE and RMSE, lower is better, while for other metrics, higher is better.}  
\label{tab:main result}%
\end{table*}%

\subsection{Overall Result Analysis}
The result of downstream tasks is presented in \tableautorefname ~\ref{tab:main result}, demonstrating that \name significantly improved the performance of all baselines across all datasets.
Skip-Gram and POI2Vec incorporate spatial information differently: Skip-Gram uses co-occurrence frequencies, while POI2Vec employs a geographic binary tree, both ignoring temporal features. Geo-Teaser includes spatial and temporal data with coarse granularity, while TALE, Hier, and CTLE integrate finer-grained spatiotemporal data. However, all six methods overlook POI semantic knowledge. Our framework addresses this gap, significantly enhancing performance. 

In three tasks, POI-Enhancer shows the most significant improvement in the Check-in Sequence Classification. This could be because the textual knowledge provided by POI-Enhancer is more beneficial for handling classification tasks. 
For the first task, POI2Vec achieves the greatest improvement on the New York dataset, with both metrics increasing by over 20\%. This is due to its focus on capturing check-in sequence patterns while neglecting other modalities. Our framework compensates for these limitations by enriching textual knowledge.
As for the second task, our findings indicate that Skip-Gram shows the weakest improvement, which is because it focuses on modeling representations from user trajectories and limits the potential for improvement.
In the last task, CTLE shows strong performance after enhancement. CTLE effectively captures contextual neighbors and temporal information in trajectories, and when combined with the textual vector extracted by POI-Enhancer, it greatly improves the performance in this tasks.

Besides, comparison experiments with LLM-based baselines reveal that, with the aid of \name, the POI representation method still holds a considerable advantage. This advantage stems from the fact that the POI representation method captures the fundamental spatial-temporal features, and when further enhanced with textual knowledge, it outperforms the text-centric LLM-based baselines. The results of this experiment are in the Supplementary Material.

\subsection{Further Analysis on \name}
\paratitle{Ablation Experiment} 
In this subsection, we conduct comprehensive experiments with four variant settings to evaluate the effectiveness of the components we design:

\bitem{POI-Enhancer/P} We remove the special prompt design including the role-playing, the attribute headers, and the question.
\bitem{POI-Enhancer/D} We removed the Dual Feature Alignment and Semantic Embedding Fusion. Instead, we generated a single prompt, which accumulates the content of the previous three kinds of prompts while maintaining the same format. The features extracted from this prompt by the LLM will be directly input into the Cross Attention Fusion.
\bitem{POI-Enhancer/F} We remove Cross Attention Fusion and concatenate the $E^{POI}$ and $E^{LLM}$ as the final vector instead.
\bitem{POI-Enhancer/C}  We only consider the spatial perspective. Specifically, given a POI, we define a square area centered around it to collect positive samples, with the parameters consistent with Geography Contrastive Learning.

\begin{figure}[t]
    \centering
    \includegraphics[width=\columnwidth]{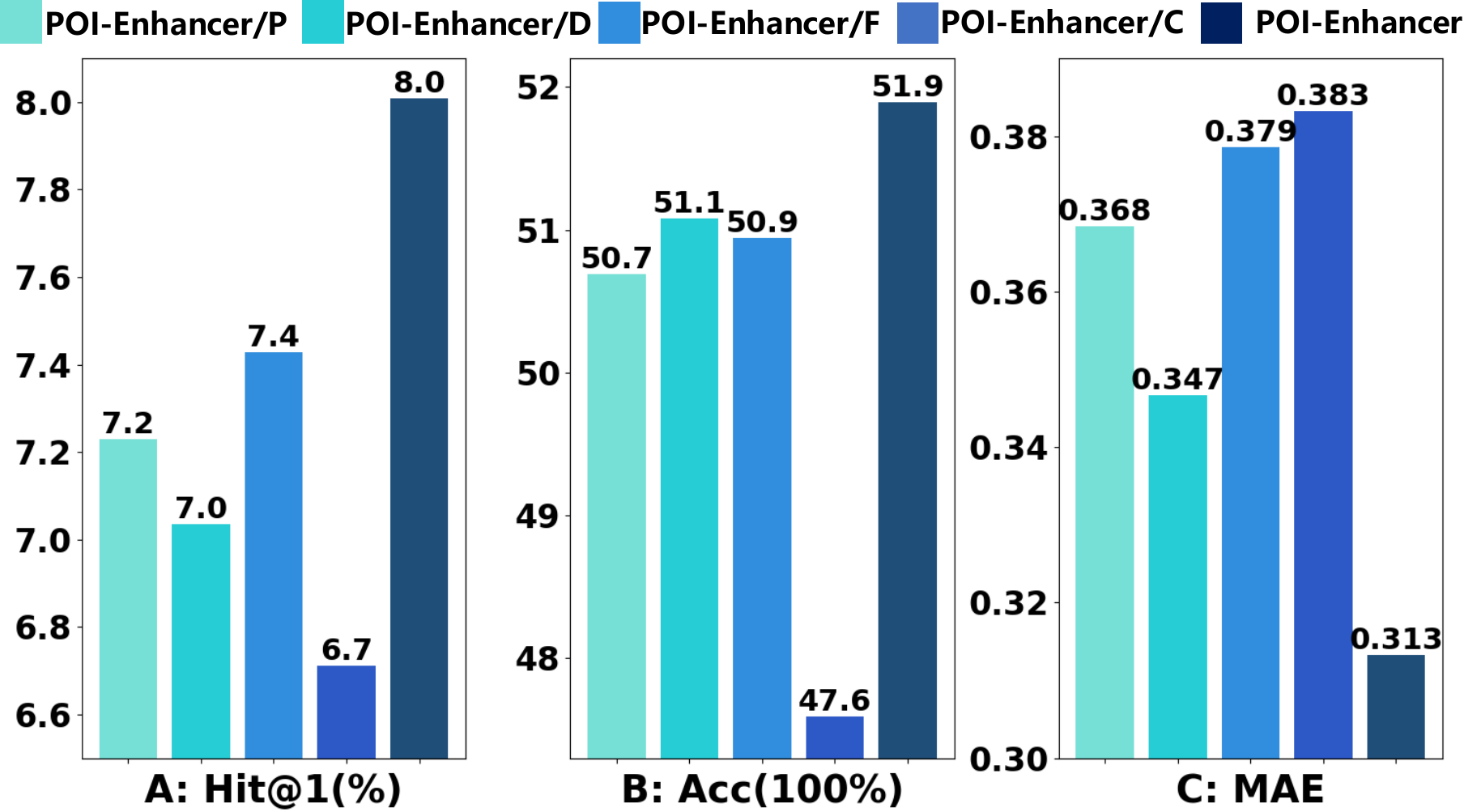}
    \caption{The result of ablation experiment. (A) is for POI Recommedation, (B) is for Check-in Sequence Classification and (C) is for POI Vistor Flow Prediction.}
    \label{fig:ablation}
\end{figure}

We tested them on three downstream tasks using the New York dataset, with Hit@1, ACC, and MAE as evaluation metrics. 
 As shown in the \figureautorefname ~\ref{fig:ablation}, POI-Enhancer outperforms all variant settings and we can draw the following conclusions:
(1) The specialized prompts can enhance the framework's performance because they stimulate the LLM to extract spatial-temporal knowledge more efficiently.
(2) The Dual Feature Alignment and the Semantic Feature Fusion help obtain and maintain high-quality semantic vectors and improve the capabilities of the POI representation.
(3) The Cross Attention Fusion enables a more thorough integration, allowing the final vector to capture richer semantic information, resulting in improved performance.
(4) Compared to distanced-based positive samples, Multi-View Contrastive Learning selects richer samples from different perspectives, enhancing the ability of the embedding vectors.

\begin{figure}[t]
    \centering
    \includegraphics[width=\columnwidth]{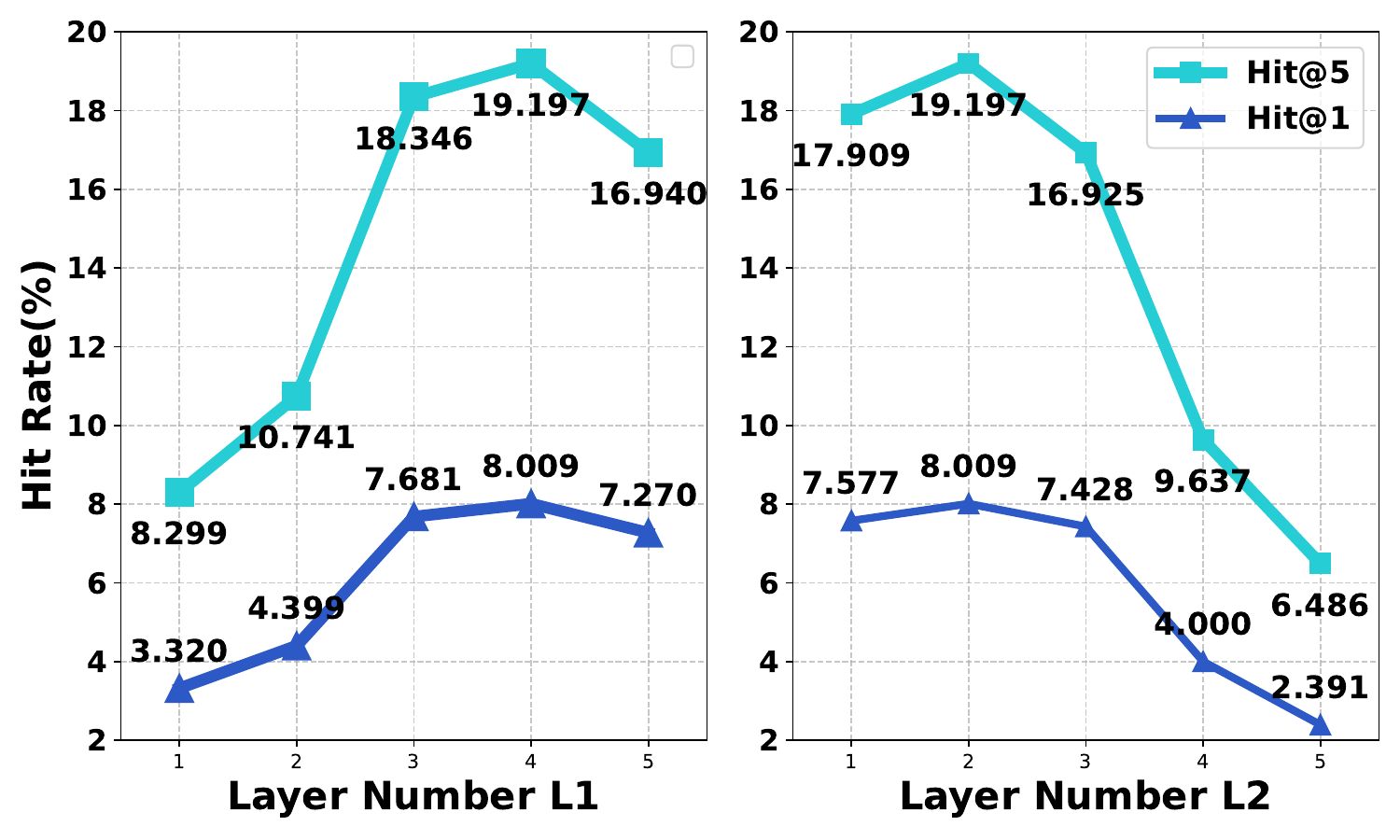}
    \caption{The effect of $L_1$ and $L_2$.}\label{fig:paprameter} 
\end{figure}

\paratitle{Parameters Analysis}
In this subsection, we study the effect of different $L_1$ and $L_2$ parameter settings in our framework. Specifically, we focus on enhancing the Hier model using the New York dataset, with POI recommendation as the downstream task. When evaluating the impact of one parameter, we keep the other parameters fixed at their optimal values.
As shown in the \figureautorefname ~\ref{fig:paprameter}, we can observe that for both $L_1$ and $L_2$, the performance initially improves with the increasing number of layers, reaches optimal performance, and then deteriorates. So, in other experiments, we set $L_1$ to 4 and $L_2$ to 2.
On the one hand, this indicates that when $L_1$ is too low, our alignment method fails to fully utilize the relational information between features, while an excessively high number of $L_1$ layers tends to cause over-fitting. On the other hand, this suggests that when $L_2$ is below the optimal value, our fusion method fails to effectively incorporate the knowledge from LLM into the original representations. However, when $L_2$ exceeds a certain threshold, the semantic knowledge will overshadow the original vectors.

\begin{figure}[t]
    \centering
    \includegraphics[width=\columnwidth]{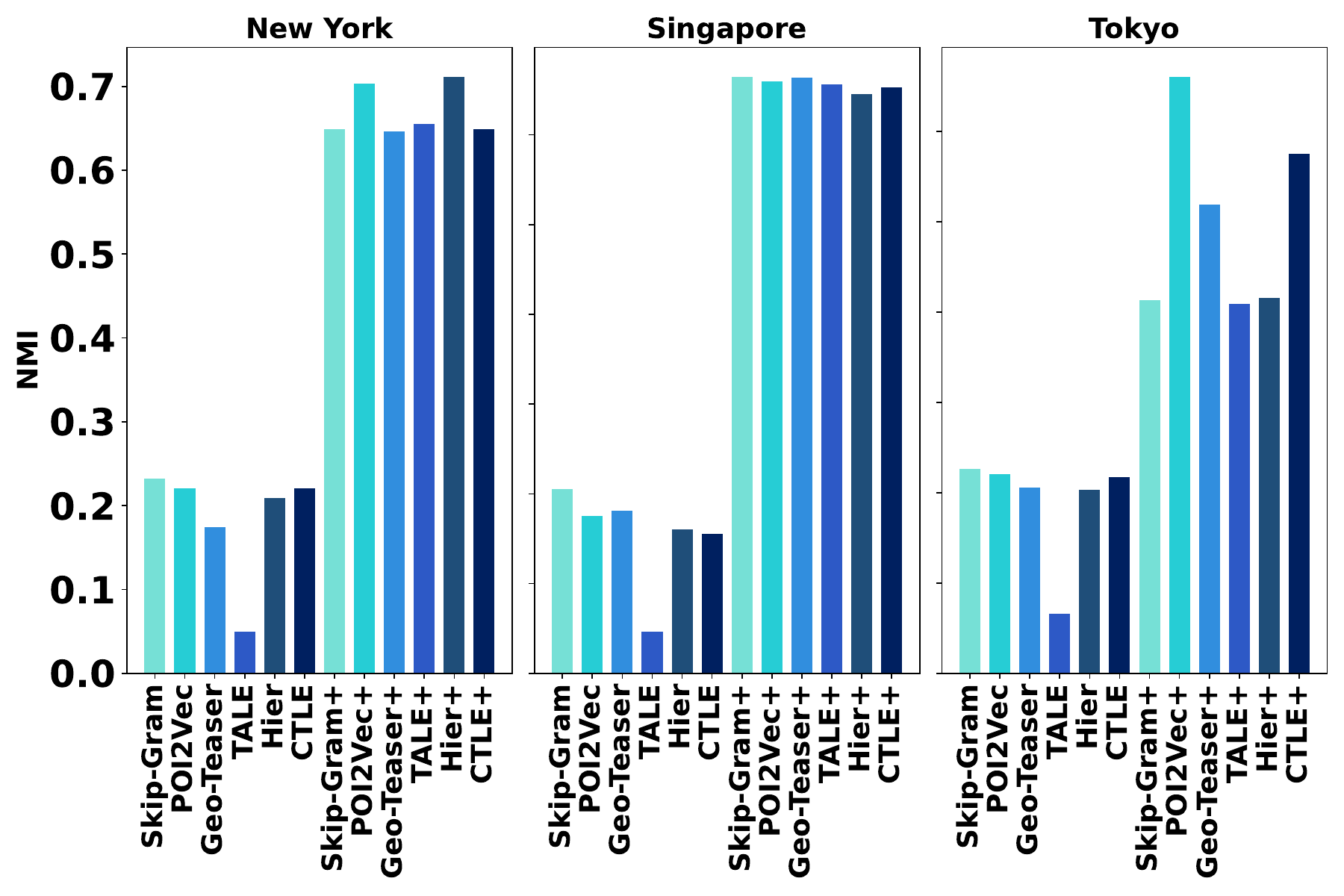}
    \caption{The result of POI cluster task.}\label{fig:cluster} 
\end{figure}

\paratitle{Quality Analysis}
To further evaluate the quality of the enhanced vectors produced by \name, we conducted clustering tasks using the K-means algorithm on three datasets. We applied this algorithm to all types of representation vectors, both before and after enhancement. The number of clusters was set to match the number of POI categories in each dataset. We then assessed the clustering performance using the Normalized Mutual Information (NMI) metric. 
The results depicted in the \figureautorefname ~\ref{fig:cluster} demonstrate the effectiveness of our framework, as all evaluation metrics for the representation vectors across the three datasets have shown significant improvement. This indicates that:
(1) We successfully extracted high-quality textual features, and the rich textual information helps similar representation vectors to cluster more closely together.
(2) We effectively integrated textual information into the initial representations, further enhancing the quality of the original vectors.
(3) The Multi-View Contrastive Learning approach encouraged vectors of the same class to be closer together while pushing vectors of different classes further apart.

\section{Related Work}\label{sec:related}
\paratitle{LLMs in Spatial-temporal Tasks}
Considerable efforts have been dedicated to using LLMs to improve the performance of spatial-temporal tasks~\cite{yu2024bigcity}. For instance, GeoGPT~\cite{geogpt} introduced an LLM-based tool capable of automating the processing of geographic data, but it does not delve into extracting detailed information about locations. GEOLLM~\cite{geollm} designed prompts that include coordinates, address, and surrounding buildings, but it can only address simple questions in a Q\&A format and are unable to handle complex tasks like POI recommendation. Besides, they fail to fully extract the semantic information of POIs.
Some researchers have used LLMs as backbones to tackle complex real-world tasks. For example, GATGPT~\cite{gatgpt} input spatial-temporal features into a frozen LLM to predict traffic speeds, while ST-LLM~\cite{stllm} used a partially frozen LLM to forecast traffic flow. However, these methods are designed for specific individual problems and cannot be applied across multiple tasks.
To solve these limitations, we designed three types of special prompts to extract the semantic information of POIs from LLMs effectively.

\paratitle{POI Representation with Semantic Information}
POI representation aims to turn each POI into a vector that can be utilized in various downstream tasks like traffic forecasting tasks~\cite{MLPST, PDFormer,stssl} and trajectory tasks~\cite{controltraj, START, gan}. Most existing methods like ~\cite{ppr}, leverage textual features typically using one-hot code to encode POI categories and then concatenate them with the embedding vectors. For data types like check-in content,~\cite{ge} model the similarity between POIs by constructing a POI-Word relationship graph, while ~\cite{cape} draws inspiration from Word2Vec method, simultaneously training word vectors and POI vectors. However, these methods often fall short in preserving semantic information and achieving a more comprehensive integration during the fusion process.
To address this issue, we designed three modules in the Embedding Enhancement to improve the preservation and integration of semantic information in the POI embedding.

\section{Conclusion and Future Work}\label{sec:conclusion}
We propose a framework called POI-Enhancer, which enhances all POI representation methods by leveraging the LLM. 
To introduce textual information into POI embeddings, we designed three special prompts to extract features from the LLM. 
To use the links between address features and other features, we introduced Dual Feature Alignment and Semantic Feature Fusion, which help obtain and preserve high-quality textual features. 
To better integrate the extracted knowledge into POI representations, we further developed the Cross Attention Fusion. 
Lastly, to enhance the representation capabilities of the vectors, we proposed Multi-View Contrastive Learning, using three strategies to sample positive and negative examples. 
The experiment results demonstrate that our framework significantly improves the performance of POI representation vectors across various downstream tasks in three real-world datasets.

\clearpage
\section*{Acknowledgments}
Prof. Jingyuan Wang's work was partially supported by the National Natural Science Foundation of China (No. 72171013, 72222022, 72242101), the Special Fund for Health Development Research of Beijing (2024-2G-30121) and State Key Laboratory of Complex \& Critical Software Environment (SKLSDE-2023ZX-04).
Prof. Xiangyu Zhao's work was partially supported by Research Impact Fund (No.R1015-23), APRC - CityU New Research Initiatives (No.9610565, Start-up Grant for New Faculty of CityU), CityU - HKIDS Early Career Research Grant (No.9360163), Hong Kong ITC Innovation and Technology Fund Midstream Research Programme for Universities Project (No.ITS/034/22MS), Hong Kong Environmental and Conservation Fund (No. 88/2022), and SIRG - CityU Strategic Interdisciplinary Research Grant (No.7020046), Huawei (Huawei Innovation Research Program), Tencent (CCF-Tencent Open Fund, Tencent Rhino-Bird Focused Research Program), Ant Group (CCF-Ant Research Fund, Ant Group Research Fund), Alibaba (CCF-Alimama Tech Kangaroo Fund No. 2024002), CCF-BaiChuan-Ebtech Foundation Model Fund, and Kuaishou.

\bibliography{aaai25}

\clearpage
\section*{Technical Appendix}

\subsection{POI Attributes Preprocess} \label{sec: data preprocess}    
    \subsubsection{Visit Pattern of POI}
    We conduct a statistical analysis of the check-in data to identify the visit patterns of a POI, which consists of the weekly visit pattern and daily visit pattern.  On the one hand, we first divide the week into weekdays (Monday to Friday) and weekends (Saturday and Sunday). Whether a POI is visited more during the weekdays or weekends will determine its weekly visit pattern. On the other hand, We part every day into seven time periods: early morning (6 to 9 AM), morning (9 to 11 AM), noon (11 AM to 1 PM), afternoon (1 PM to 5 PM), evening (5 to 7 PM), night (7 to 12 PM), and midnight (0 to 6 AM).  Afterward, we categorize each POI's check-in records into the corresponding time slot. Finally, the time slot with the highest number of check-in records will represent the daily visit pattern. Take the POI from \tableautorefname~\ref{tab:poi_example} as an example: its weekly visit pattern is weekdays, and its daily visit pattern is 6 to 9 AM.

    \subsubsection{Address of POI}
    Address features are essential for a POI. Typically, a POI's address includes the street name, house number, and postal code.  By utilizing geographic revere search API provided by Nominatim \footnote{Nominatim: \url{https://nominatim.openstreetmap.org}}, we can input latitude and longitude coordinates to retrieve the corresponding address details.  For each POI in the dataset, we leverage this API to query its address information as well as the basic attribute $Name$. 

    \subsubsection{Surrounding of POI}
    For any given POI, we survey the category of other POIs nearby and consider this as the surrounding attributes. Specifically,  we search for all POIs within a square area centered on a certain POI and the square's side length is set to 0.5 km. Then, we count the number of categories within these POIs and sort them in descending order. Finally, we select the top three categories as its surrounding attribute. For example, the top three categories near the POI in \tableautorefname~\ref{tab:poi_example} are office, building, and road, which are its surrounding attributes.

\subsection{Multi-View Contrastive Learning}

\paratitle{Sequence-Time Contrastive Learning}
    A formal constraint is that given a check-in record sequence $R$ and a check-in record $r=(u,p,t)$, for any $r'=(u,p',t')$ in the $R$, we have the following two requirements for $r'$:
    \begin{itemize}
        \item $|index(r') - index(r)| \leq \lambda$, where $index$ is a function indicating the position in $R$, and $\lambda$ is a threshold.
        \item $date(r') = date(r)$, where $date$ is a function that extracts the date from a timestamp.
    \end{itemize}

    The $p'$ in $r'$ which meets the above requirements is considered a positive sample.  

\paratitle{Geography Contrastive Learning}
    Given a POI $p$, the positive samples $p'$ have to meet the following two criteria:
    \begin{itemize}
        \item $Area(p)$ is defined as $Square(p, \tau)$, where $Square$ is a function that generates the corresponding square region with a side length of $\tau$. The point $p'$ lies within the $Area(p)$.
        \item $p'.c=p.c$, and $c$ is the category attribute of a POI introduced in Section ~\ref{sec:pre}.
    \end{itemize}

\paratitle{Functional Contrastive Learning}
Given a POI $p$, the positive samples $p'$ must satisfy the following two conditions:
    \begin{itemize}
        \item $p'.visit \_pattern=p.visit\_pattern$, and $visit\_pattern$ is the visit pattern attribute of a POI.
        \item $p'.c=p.c$, where $c$ is the category attribute of a POI. 
    \end{itemize}

\subsection{Experiments}
In our experiments section, the training processes of the \name and all downstream tasks are implemented with Pytorch on the Nvidia RTX 3090 GPU. For LLM-based baselines, We use the POI Address Prompts to extract features from the LLM as representations.

\paratitle{Baselines}

\bitem{Skip-Gram}~\cite{mikolov2013efficient}, a simple variant of Word2Vec model, and is widely used in sequential tasks.

\bitem{POI2Vec}~\cite{feng2017poi2vec}, a latent representation model for both POI and user, which introduces geographical information into output via splitting the map into rectangle regions and building binary tree on it.

\bitem{Geo-Teaser}~\cite{zhao2017geo}, an embedding model based on Skip-Gram, but take temporal influence and neighboring locations in the trajectory dataset into consideration and design two separate loss functions for them.

\bitem{TALE}~\cite{wan2021pre}, a POI embedding pre-training method based on CBOW,  constructing a temporal tree structure based on user trajectories to acquire time awareness.

\bitem{Hier}~\cite{shimizu2020enabling}, a method to obtain fine-grained place embedding via extracting spatial information from trajectory dataset in a hierarchical way.

\bitem{CTLE}~\cite{lin2021pre}, the state-of-the-art model of POI representation which incorporates both temporal and spatial information of user activities into its embedding.

\paratitle{Statistics of Processed Datasets}
After processing the data, the details of the datasets are presented in \tableautorefname\ref{tab:dataset_sta}.
\begin{table}[t]
\centering
    \resizebox{0.7\columnwidth}{!}{%
        \begin{tabular}{rccc}
            \toprule
            Dataset & \#User & \#POI & \#Check-in \\
            \midrule
            Foursquare-NY & 15,171 & 24,118 & 641,005 \\ 
            Foursquare-SG & 10,909 & 20,154 & 696,306 \\ 
            Foursquare-TKY & 2,293 & 15,164 & 496,459 \\
            \bottomrule
        \end{tabular}
    }
     \caption{Statistics of Datasets.} 
    \label{tab:dataset_sta}
\end{table}

\paratitle{Downstream Task Implementation}
In our framework, the dimension $d$ is uniformly set to 256.  Moreover, the number of encoder layers $L_1$ in Dual Feature Alignment and the number of encoder layers $L_2$ in Cross Attention Fusion is 4 and 2, respectively. The temperature parameter $\gamma$ in InfoNCE is set to 0.1. We train \name for 100 epochs using a learning rate of 0.001 and the AdamW optimizer with a decay of 0.001. 

\bitem{POI Recommendation},
To evaluate this task, we train a two-layer LSTM model to process the input check-in sequence and feed the output into an MLP to make the prediction. Note that we cut long sequences in the dataset into slices with less than 128 check-in records to improve model efficiency.

\bitem{Check-in Sequence Classification},
Similar to the POI Recommendation task, we use a two-layer LSTM connected to an MLP to classify check-in sequences.

\bitem{POI Visitor Flow Prediction},
 We set the time window to 1 hour and calculate the check-in count of every POI based on the check-in dataset. Due to the sparsity of the check-in dataset, we only select non-zero visitor flow sequences with a length greater than 5 when building the dataset. Then, we trained a Seq2seq model with MSE loss to do the prediction. Before input into the model, the visitor flow series is normalized and appended with two additional information, the embedding of the current POI and the hour of the day at which the series starts.

\paratitle{Parameter Settings}

In \name, the distance $\tau$ is set to 0.5 km when obtaining POI Surrounding features. The feature dimension $D$ extracted from the Llama2 and ChatGLM2 is 4096, while the feature dimension $D$ extracted from GPT-2 is 768.
Within each encoder layer of the Dual Feature Alignment and Cross Attention Fusion, the number of attention heads $H$ is set to 8, and the dimension of each head $d_h$ is set to 32.

Besides,  $\lambda$ in Sequence-Time Contrastive Learning is set to 2 and the sampling distance of Geography Contrastive Learning $\tau$ is set to 0.5 km.

 In the downstream task training, the hidden size of LSTM is uniformly set to 512. We use the Adam optimizer and a learning rate of 0.0001 to train all the downstream models for 100 epochs, except for the model of the POI Recommendation task, which has a learning rate of 0.001.

\subsection{Case Study}
Here, we conduct distance comparison on the POI  representation vectors before and after enhancement. Specifically,  we selected two POIs that are geographically very close and of the same category from the New York dataset. Despite the high similarity between these two POIs, through the output of the LLM, we can identify disparities in their semantic information. For specific attributes, please refer to the \tableautorefname ~\ref{tab: similar_POI}.
    \begin{table}[ht]
    \centering
        \resizebox{\columnwidth}{!}{%
            \begin{tabular}{rccc}
                \toprule
                POI Name & \#Long, Lat & \#Category & \#LLM Output \\
                \midrule
                Bagatelle & -74.006,40.739 & French Restaurant & festive atmosphere \& brunches \\ 
                The Jane & -74.009,40.738 & French Restaurant & cozy ambiance \& cocktails \\ 
                \bottomrule
            \end{tabular}
        }
         \caption{The attributes of two similar POIs.} 
        \label{tab: similar_POI}
    \end{table}
Subsequently, we utilized the Hier model before and after enhancement to represent these two POIs. The Euclidean distance was employed to calculate the distance between the resulting representation vectors, and the obtained results are presented in \tableautorefname ~\ref{tab: distance_POI}. We find the Euclidean distance between POI vectors was smaller in Hier, while it was larger in Hier+, showing POI-Enhancer can enrich the vectors with LLM knowledge.

\begin{table}[htbp]
    \centering
        \resizebox{0.4\columnwidth}{!}{%
            \begin{tabular}{rc}
                \toprule
                Model & Distance \\
                \midrule
                Hier+ &   262.90 \\
                \midrule
                Hier & 7.49 \\
                \bottomrule
            \end{tabular}
        }
         \caption{The distance comparison result.}
        \label{tab: distance_POI}
    \end{table}

    \subsection{Performance of LLM-based baselines}

    \begin{table}[htbp]
\belowrulesep=0pt
\aboverulesep=0pt
\setlength{\tabcolsep}{1.2mm}
\centering
\resizebox{\columnwidth}{!}{
    \begin{tabular}{l|cc|cc|cc}
    \toprule
    \textbf{Task} & \multicolumn{6}{c|}{\textbf{POI Recommendation}}  \\
    \midrule
    Dataset & \multicolumn{2}{c|}{NY} & \multicolumn{2}{c|}{TKY} & \multicolumn{2}{c|}{SG}\\
    \midrule
    Metric & Hit@1 & Hit@5 & Hit@1 & Hit@5 & Hit@1 & Hit@5   \\
    \midrule
    Skip-Gram+ & 7.610  & 18.032  & \textcolor[rgb]{ .122,  .137,  .161}{15.557 } & \textcolor[rgb]{ .122,  .137,  .161}{34.197 } & \textcolor[rgb]{ .122,  .137,  .161}{10.747 } & \textcolor[rgb]{ .122,  .137,  .161}{24.468 }   \\
    POI2Vec+ & 7.851  & 18.353  & 15.800  & 34.768  & 10.630  & 24.030    \\
    Geo-Teaser+ & \textcolor[rgb]{ .216,  .235,  .263}{7.116 } & \textcolor[rgb]{ .216,  .235,  .263}{16.657 } & 15.500  & 34.475  & \textcolor[rgb]{ .122,  .137,  .161}{10.122 } & \textcolor[rgb]{ .122,  .137,  .161}{23.532 }   \\
    TALE+ & 6.690  & 15.208  & 14.940  & 33.223  & 8.694  & 20.342   \\
    Hier+ & 8.009  & 19.197  & 16.187  & 35.715  & \textcolor[rgb]{ .122,  .137,  .161}{10.592 } & \textcolor[rgb]{ .122,  .137,  .161}{24.079 }   \\
    CTLE+ & \textcolor[rgb]{ .216,  .235,  .263}{7.093 } & \textcolor[rgb]{ .216,  .235,  .263}{17.032 } & \textcolor[rgb]{ .216,  .235,  .263}{15.479 } & 34.138  & \textcolor[rgb]{ .122,  .137,  .161}{10.315 } & \textcolor[rgb]{ .122,  .137,  .161}{24.027 }   \\
    \midrule
    GPT-2 & 0.6639  & 2.7265  & 6.9243  & 17.3328  & 2.3632  & 8.0120    \\
    Llama2 & 1.6491  & 5.1294  & 8.0065  & 19.0356  & 3.9478  & 11.8564    \\
    ChatGLM2 & 4.2386  & 11.4755  & 12.4734  & 28.2362  & 5.4692  & 16.1045   \\
    \end{tabular}%
    }
    \caption{The performance of LLM-based baselines in downstream tasks.}
  \label{tab:LLM-based baseline1}
  \end{table}

  \begin{table}[htbp]
\belowrulesep=0pt
\aboverulesep=0pt
\setlength{\tabcolsep}{1.2mm}
\centering
\resizebox{\columnwidth}{!}{
    \begin{tabular}{l|cc|cc|cc}
    \toprule
    \textbf{Task} & \multicolumn{6}{c|}{\textbf{Check-in Sequence Classification}} \\
    \midrule
    Dataset & \multicolumn{2}{c|}{NY} & \multicolumn{2}{c|}{TKY} & \multicolumn{2}{c|}{SG} \\
    \midrule
    Metric  Acc   & F1    & Acc   & F1    & Acc   & F1    \\
    \midrule
    Skip-Gram+  & 52.151  & 0.251  & 62.936  & 0.438  & \textcolor[rgb]{ .122,  .137,  .161}{47.285 } & 0.255   \\
    POI2Vec+ & 52.151  & 0.245  & 62.358  & 0.438  & 46.521  & 0.264   \\
    Geo-Teaser+ & 49.910  & 0.233  & 62.647  & 0.437  & 50.064  & 0.279    \\
    TALE+   & 50.689  & 0.240  & 63.380  & 0.448  & 47.719  & 0.263  \\
    Hier+ &  51.893  & 0.254  & 63.380  & 0.441  & 47.795  & 0.258    \\
    CTLE+ &  50.430  & 0.234  & 61.848  & 0.434  & 51.440  & 0.287    \\
    \midrule
    GPT-2 &  1.0327  & 0.0006  & 0.4664  & 0.0003  & 0.6628  & 0.0004    \\
    Llama2 & 1.0327  & 0.0005  & 1.3324  & 0.0039  & 0.9177  & 0.0019   \\
    ChatGLM2  & 17.1687  & 0.0487  & 31.3347  & 0.1680  & 11.5728  & 0.0372    \\
    \end{tabular}%
    }
    \caption{The performance of LLM-based baselines in Check-in Sequence Classification.}
  \label{tab:LLM-based baseline2}
  \end{table}

\begin{table}[htbp]
\belowrulesep=0pt
\aboverulesep=0pt
\setlength{\tabcolsep}{1.2mm}
\centering
\resizebox{\linewidth}{!}{
    \begin{tabular}{l|cc|cc|cc}
    \toprule
    \textbf{Task} &  \multicolumn{6}{c|}{\textbf{POI Visitors Flow Prediction}} \\
    \midrule
    Dataset & \multicolumn{2}{c|}{NY} & \multicolumn{2}{c|}{TKY} & \multicolumn{2}{c|}{SG}  \\
    \midrule
    Metric   & MAE   & RMSE  & MAE   & RMSE  & MAE   &  RMSE  \\
    \midrule
    Skip-Gram+   & 0.336  & 0.514  & 0.492  & 0.668  & 0.621  & 0.890  \\
    POI2Vec+ & 0.326  & 0.492  & 0.490  & 0.696  & 0.602  & 0.868  \\
    Geo-Teaser+  & 0.341  & 0.524  & \textcolor[rgb]{ .122,  .137,  .161}{0.483 } & \textcolor[rgb]{ .122,  .137,  .161}{0.669 } & 0.588  & 0.854  \\
    TALE+ & 0.320  & 0.482  & 0.510  & 0.701  & \textcolor[rgb]{ .122,  .137,  .161}{0.610 } & \textcolor[rgb]{ .122,  .137,  .161}{0.903 } \\
    Hier+ &  0.313  & 0.483  & \textcolor[rgb]{ .122,  .137,  .161}{0.510 } & \textcolor[rgb]{ .122,  .137,  .161}{0.719 } & 0.574  & 0.804  \\
    CTLE+ &  0.291  & 0.456  & \textcolor[rgb]{ .122,  .137,  .161}{0.495 } & \textcolor[rgb]{ .122,  .137,  .161}{0.689 } & 0.610  & 0.892  \\
    \midrule
    GPT-2 & 0.409  & 0.773  & 0.539  & 0.766  & 0.637  & 0.923  \\
    Llama2 & 0.389  & 0.694  & 0.520  & 0.738  & 0.639  & 0.901  \\
    ChatGLM2 & 0.439  & 0.795  & 0.536  & 0.738  & 0.629  & 0.952  \\
    \end{tabular}%
    }
    \caption{The performance of LLM-based baselines in POI Visitors Flow Prediction.}
  \label{tab:LLM-based baseline3}%
  \end{table}

 Here, we utilize three LLM-based baselines including Llama2, ChatGLM2, and GPT-2. Specifically, we used the POI Address Prompts to extract features from the LLM as representations, just as we discussed in Section ~\ref{sec:method}. Then, we compare the LLM-based baselines with the POI embedding models improved by \name. The result of this experiment is shown in \tableautorefname ~\ref{tab:LLM-based baseline1}, \tableautorefname ~\ref{tab:LLM-based baseline2}, \tableautorefname ~\ref{tab:LLM-based baseline3}.

   Furthermore, to examine the contribution of the LLM and Contrastive Learning, we design two experiments on the NY dataset, using the Hier model and the POI recommendation task. The details of the experiment settings are shown below. 

   \begin{table}[htbp]
    \centering
        \resizebox{0.7\columnwidth}{!}{%
            \begin{tabular}{rcc}
                \toprule
                Model & Hit@1 & Hit@5 \\
                \midrule
                Hier+ & 8.009 & 19.197 \\ 
                \midrule
                Hier+(w/o LLM)& 7.029 & 17.028 \\
                \midrule
                Hier+(w/o CL)& 8.009 & 19.197 \\
                \bottomrule
            \end{tabular}
        }
         \caption{The attributes of two similar POIs.} 
        \label{tab:additional_ablation}
    \end{table}
    
    \bitem{Hier+} We use the Hier model enhanced by our framework.
  
    \bitem{Hier+(w/o LLM)} We remove the special prompt design including the role-playing, the attribute headers, and the question.
    
    \bitem{Hier+(w/o CL)} We replaced the fusion process with vector addition and removed the Contrastive Learning Scheme from the framework.

    This section ablation experiment result in the \tableautorefname ~\ref{tab:additional_ablation} shows that LLM and Contrastive Learning both are beneficial to the performance.

\begin{figure}[h]
    \centering
    \includegraphics[width=0.4\textwidth]{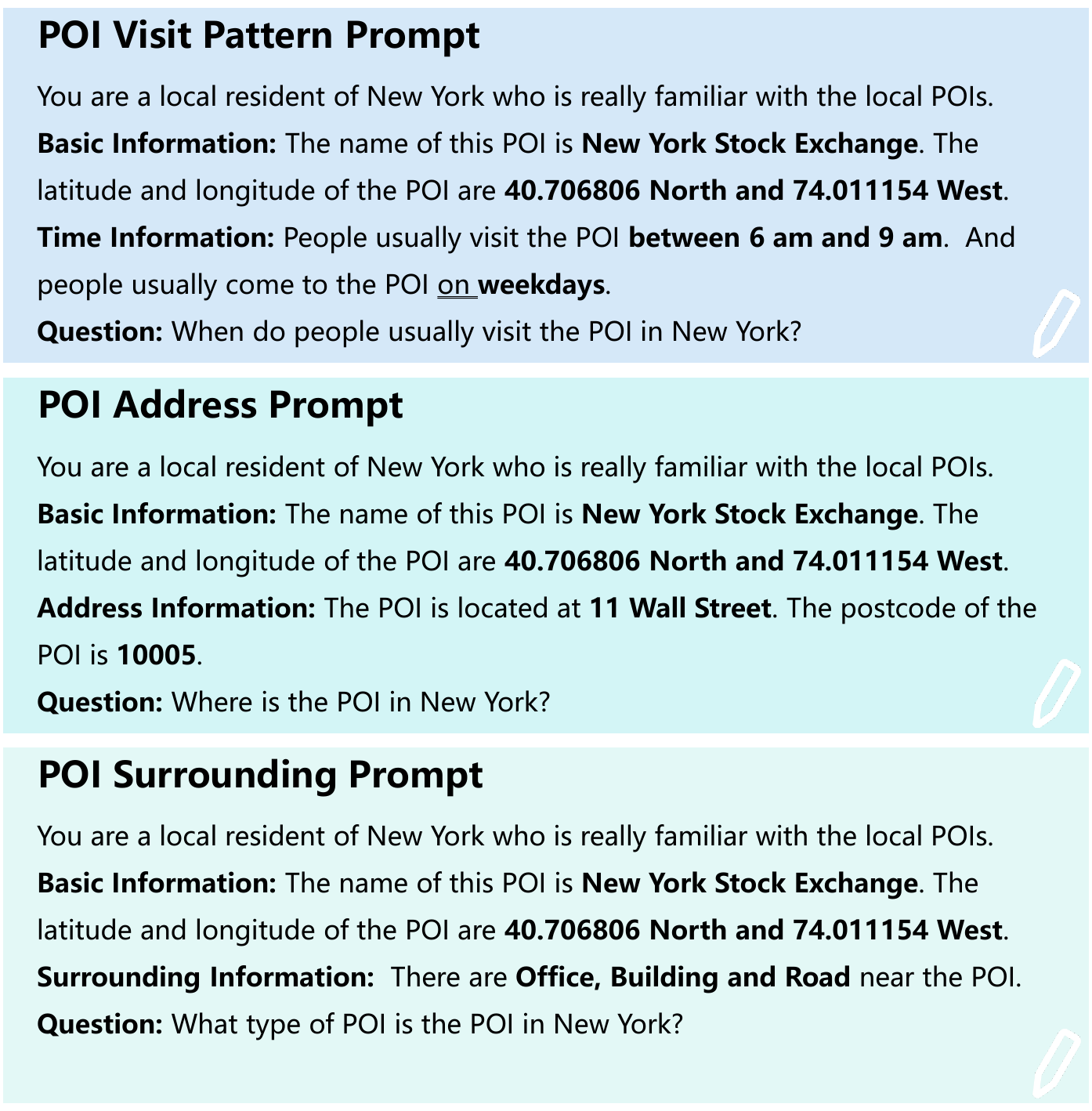} 
    \caption{The examples of special prompt we design}.  
    \label{fig: prompt}
\end{figure}

\end{document}